\documentclass[10pt,twocolumn,letterpaper]{article}

\usepackage{cvpr}
\usepackage{times}
\usepackage{epsfig}
\usepackage{graphicx}
\usepackage{amsmath}
\usepackage{amssymb}

\usepackage{multirow}
\usepackage{caption}
\usepackage{subcaption}
\usepackage{color}

\usepackage[pagebackref=true,breaklinks=true,letterpaper=true,colorlinks,bookmarks=false]{hyperref}

\cvprfinalcopy % *** Uncomment this line for the final submission

\begin{document}

%%%%%%%%% TITLE
\title{Scalable and Effective Deep CCA via Soft Decorrelation}

\author{Xiaobin Chang$^1$, Tao Xiang$^1$, Timothy M. Hospedales$^2$\\
Queen Mary University of London$^1$, The University of Edinburgh$^2$\\
{\tt\small \{x.chang, t.xiang\}@qmul.ac.uk t.hospedales@ed.ac.uk}
%\and
}

\maketitle
%\thispagestyle{empty}

%%%%%%%%% ABSTRACT
\begin{abstract}
Recently the widely used multi-view learning model, Canonical Correlation Analysis (CCA) has been generalised to the non-linear setting via deep neural networks.
Existing deep CCA models typically first decorrelate the feature dimensions of each view before the different views are maximally correlated in a common latent space. 
This feature decorrelation is achieved by enforcing an exact decorrelation constraint; these models are thus computationally expensive due to the matrix inversion or SVD operations required for exact decorrelation at each training iteration.
%\textcolor{black}{Existing deep CCA models handle decorrelation constraints by using exact decorrelation which is computationally expensive due to the matrix inversion or SVD operations are required at each training iteration.}
% Taking an exact or `hard' decorrelation step, these models are computationally expensive due to the necessary matrix inversion or SVD decomposition operations in each training iteration. 
Furthermore, the decorrelation step is often separated from the gradient descent based optimisation, resulting in sub-optimal solutions. We propose a novel deep CCA model \emph{Soft~CCA} to overcome these problems. Specifically, exact decorrelation is replaced by soft decorrelation via a mini-batch based Stochastic Decorrelation Loss (SDL) to be optimised jointly with the other training objectives. Extensive experiments show that the proposed soft~CCA is more effective and efficient than existing deep CCA models. In addition, our SDL loss can be applied to other deep models beyond multi-view learning,
%where feature decorrelation is beneficial, 
and obtains superior performance compared to existing decorrelation losses.
\end{abstract}

%%%%%%%%% BODY TEXT
\section{Introduction}

Canonical Correlation Analysis (CCA)~\cite{Basic_CCA,cca_theory} is widely used for multi-view learning.  These views could be camera views, e.g., the images of a face from different view angles,  or modalities, e.g., an image and its caption.  CCA aims to learn a joint embedding space where different views of a single data item are maximally correlated/aligned. Many tasks can be accomplished in this space such as cross-view recognition, retrieval and synthesis \cite{multiview_CCA,Zhu_NIPS2014,KanCVPR16,DCCA_full_batch,DCCA_large_batch,CorrNet}.

A standard CCA model is linear in the sense that the projection between the feature space and the embedding space is linear. For learning richer non-linear embeddings, Kernel CCA (KCCA)~\cite{KCCA} extended linear CCA via kernelisation. Both linear CCA and KCCA are shallow models and the training procedure usually requires accessing the whole batch data.
As a result, KCCA has poor scalability.
The recently proposed deep CCA ~\cite{DCCA_full_batch, DCCA_large_batch, ImgTxt_DCCA,wangSDCCA2015} aims to learn nonlinear projections with deep neural networks rather than kernels and has been shown to be more effective than shallow CCA and KCCA. 

However,  scalability issues remain for deep CCA. This is because existing deep CCA models \cite{DCCA_full_batch,DCCA_large_batch,ImgTxt_DCCA,wangSDCCA2015} aim to implement an exact or `hard' decorrelation. More precisely, before being projected into the common embedding space, the extracted deep feature vector for each view is decorrelated by forcing its correlation matrix over the training batch to be an identity matrix. This decorrelation operation is exact but computationally expensive. Either matrix inversion~\cite{DCCA_full_batch,DCCA_large_batch} or singular value decomposition (SVD)~\cite{wangSDCCA2015} is required \emph{at each iteration} which severely limits scalability. Furthermore, existing deep CCA models such as ~\cite{wangSDCCA2015} typically employ two separate and independent optimisation steps:  
%exact decorrelation described above, followed by gradient computation and subsequent back propagation~\cite{wangSDCCA2015}.
%Without joint optimisation, this could lead to sub-optimal solutions. 
\textcolor{black}{The feature representation for each data view is first decorrelated exactly as described above. These decorrelation operations do not directly affect the following gradient computation and subsequent backpropagation.
Without jointly optimising the decorrelation constraint and other learning objectives, this could lead to sub-optimal solutions. } 

In this paper, we propose Soft CCA, a novel approach to deep CCA.  In our model,  decorrelation is formulated as a soft constraint to be jointly optimised with other training objectives. Specifically,  a robust decorrelation loss, called Stochastic Decorrelation Loss (SDL), is introduced, which is mini-batch based and approximates the full-batch statistics efficiently and effectively by using stochastic incremental learning.  SDL  is a softer constraint as the loss is only minimised rather than enforced to be zero. Comparing with existing deep CCA models, Soft CCA has two advantages:  First, it is more efficient and scalable -- by avoiding computationally expensive operations such as SVD, its cost is quadratic $O(k^2)$ rather than cubic $O(k^3)$ with a $k$-dimensional feature input. Second, by jointly optimising the decorrelation loss with other losses such as the distance between views in the embedding space, more globally optimal solutions can be achieved resulting in more effective correlation analysis and learning of multi-view embeddings.  

%As a decorrelation technique, the proposed SDL can also be applied to any deep model where feature decorrelation is helpful. 
%In this work, we demonstrate this with two widely used models including Factorisation Autoencoder (FAE) and CNN classification -- with SDL as an activation regularisation. 
%For the FAE model which aims to disentangle latent factors of variation that correspond to different aspects of data items, SDL is used to ensure that the representations of distinct factors are disentangled.
%For a deep classification, SDL can alleviate the overfitting and maximise capacity by whitening the computed deep features. 
\textcolor{black}{
While our proposed SDL is motivated by the feature decorrelation required for deep CCA learning, it can also be applied as an activation regularisation to any deep model where feature decorrelation is helpful.
In this work, we demonstrate this with two widely used models including Factorisation Autoencoder (FAE) and convolutional neural network (CNN) based classifiers. 
 FAE architectures aim to disentangle latent factors of variation that correspond to different aspects of data items. Here we use SDL-based decorrelation to ensure representations of distinct factors are indeed disentangled, and show that it provides superior disentangling performance compared to prior approaches.
  As for the supervised CNN classifier, it was recently shown that decorrelation losses can be beneficial for maximizing model capacity and reducing overfitting \cite{decov_l2_loss}. In this case, we show that by whitening the computed deep features in supervised CNN classifiers, we can train a more effective classifier for both instance and category-level  recognition benchmarks.}
%As a decorrelation technique, the proposed SDL can also be applied to achieve different objectives.
%In this work, we focus on two purposes. Disentangling latent factors of variation that correspond to different aspects of data items and maximising model capacity by whitening the computed deep features.
%A Factorisation Autoencoder (FAE) is used to demonstrate the first purpose while 
%demonstrate this with two widely used models including Autoencoder and CNN classification -- with SDL as an activation regularisation.
%On the one hand, two types of Autoencoder are discussed.
%First we focus on Factorisation Autoencoder (FAE) model which aims to disentangle latent factors of variation that correspond to different aspects of data items, SDL is used to ensure that the representations of distinct factors are disentangled.
%For the conventional Autoencoder (AE), the capacity of final encoded representation can be enlarged by decorrelating different feature dimensions with SDL and reconstruction quality thus can be improved.
%On the other hand, for a deep classification model, SDL can alleviate the overfitting and maximise capacity by whitening the computed deep features.

We conduct extensive experiments on multi-view correlations analysis. The results show that the proposed soft deep CCA is much more efficient as well as more effective than the existing shallow or deep CCA models -- and is also simpler to implement. Moreover, we demonstrate that SDL can be applied to a number of models for problems beyond multi-view learning, and improves model performance beyond that of existing decorrelation losses. 

\section{Related Work}\label{RelatedWork}
\subsection{Deep~CCA}
%\noindent{\textbf{Deep~CCA}} \quad 
Canonical Correlation Analysis (CCA)~\cite{Basic_CCA} and its variants including Kernel CCA~\cite{KCCA} and multi-view CCA~\cite{multiview_CCA} are one of the most popular multi-view learning approaches. Inspired by the success of Deep Neural Network (DNNs) in  representation learning~\cite{deep_feats}, Deep CCA has received increasing interest \cite{DCCA_full_batch,DCCA_large_batch,wangSDCCA2015}.
A deep CCA architecture was first proposed by Deep CCA (DCCA)~\cite{DCCA_full_batch} which directly computes the gradients of CCA objective and requires both a second-order optimisation method~\cite{L-BFGS_opts} and full-batch training inputs. It thus cannot cope with large training data sizes.
An alternative deep CCA objective and architecture are proposed in Stochastic Deep CCA (SDCCA)~\cite{wangSDCCA2015} which make it suitable for mini-batch stochastic optimisation. However, due to the exact decorrelation used, SDCCA still requires a costly SVD operation at each training iteration. SVD's $O(k^3)$ cost is not scalable to the large layer sizes $k$ (e.g., $k=1024$)  common in contemporary DNNs. In fact, all existing deep CCA models \cite{DCCA_full_batch,DCCA_large_batch,wangSDCCA2015}  take an exact decorrelation step, which limits their scalability and effectiveness as mentioned earlier. Furthermore, the exact decorrelating operations often do not directly impact the following gradient computations and backpropagation,  which could lead to sub-optimal optimisation.  
In contrast, our Soft Deep CCA decorrelates by formulating the decorrelation constraint as a loss which is optimised end-to-end jointly with other losses in a standard SGD procedure, making it both more scalable and more effective.

%\noindent{\textbf{Other Multi-view learning model}} \quad Another popular Multi-view learning model is the Multimodal Autoencoder (MAE) \cite{multimodal_AE} which aims to achieve both within-view and across view reconstruction via a shared embedding. MAEs have also been combined with CCA \cite{CorrNet,wang2015-deep-multi-view}. Apart from aligning multiple views, multi-view learning has also been optimised for supervised classification  \cite{KanCVPR16}. Soft CCA is orthogonal to these and can be readily integrated into them. 

%\noindent{\textbf{Decorrelation Loss}} \quad 
\subsection{Decorrelation Loss}
Beyond multi-view learning, many other deep models benefit from decorrelation of activations in a neural network layer. For these models, a decorrelation loss such as the proposed SDL can be employed.  
\textcolor{black}{Two such models are studied in this work, namely the Factorisation Autoencoder (FAE),  and convolutional neural network (CNN) based classifiers.}  For each model, an alternative decorrelation loss exists.
%Two examples are studied in this work, including Factorisation Autoencoder (FAE) and a deep classification convolutional neural network (CNN).  

\noindent{\textbf{FAE and XCov loss}} Recently interest has regrown in models for disentangling the underlying factors of variation in the appearance of objects in images, for example identity and viewpoint \cite{Zhu_NIPS2014,Wen_2016_CVPR,bayesian_rep_learn,
conditional_similarity,VAE,disentangling_VAE,AAE}.  FAE  achieves \textcolor{black}{semi-supervised}  disentangling of latent factors via a two-branch  autoencoder. Recently it has been shown in \cite{Discover_Hidden_Factor}   that the efficacy of FAE can be improved by adding a decorrelation loss (termed XCov in \cite{Discover_Hidden_Factor}) to explicitly decorrelate the computed latent factor representations. Like our SDL,  computing XCov is also a mini-batch operation. But it only eliminates correlations across and not within each factor; and it computes covariance only within each mini-batch, while our SDL approximates full-batch statistics using stochastic incremental learning. We show in our experiments (Sec.~\ref{sec:fae}) that SDL is more effective than XCov for helping FAE to disentangle latent factors.

\noindent{\textbf{CNN Classifier and DeCov loss}} \textcolor{black}{ Using CNN with a classification loss (e.g., cross entropy) for object recognition is perhaps the most popular application of deep learning in computer vision. CNN classifiers are used for not only object category recognition tasks \cite{CIFAR10_Data, krizhevsky2012imagenet} but also object instance/identity  recognition/verification tasks such as face verification \cite{Deep_Face_joint_NIPS2014} and person re-identification \cite{DGD}. 
When training CNNs for classification, avoiding overfitting, saturation and slow convergence are crucial~\cite{train_dnn_hard}. These problems are often alleviated by regularisation such as Batch Normalisation~\cite{batch_norm} and dropout~\cite{dropout}. Recently it was shown that decorrelation losses can also be used for effective overfitting reduction \cite{decov_l2_loss}.}
%Perhaps the most widely used deep models are  CNNs designed for classification. When training CNNs, avoiding overfitting, saturation and slow convergence are crucial~\cite{train_dnn_hard}. These problems are often alleviated by regularisation such as Batch Normalisation~\cite{batch_norm} and dropout~\cite{dropout}. Recently it was shown that decorrelation losses \textcolor{black}{can also be used for effective overfitting reduction \cite{decov_l2_loss}}.
Compared with the existing decorrelation loss DeCov~\cite{decov_l2_loss}, our SDL has the following advantages: (1) More accurate covariance statistics due to full-batch approximation instead of the pure mini-batch statistics used in DeCov~\cite{decov_l2_loss}. (2) SDL uses a more robust $L_1$ formuation instead of the $L_2$ one in DeCov~\cite{decov_l2_loss}, which encourages sparser correlation and thus stronger decorrelation.

\noindent{\textbf{Our contributions}} are as follows: (1)  We provide a new perspective on CCA that allows its objective to be expressed as a loss to be minimised by gradient descent rather than as an eigen-decomposition problem. (2) We propose Soft CCA, a novel Deep CCA model that is simple to implement, more efficient and scalable (mini-batch SGD-based optimisation) and more effective (full batch approximation, jointly end-to-end) than existing deep CCA models. (3) Beyond multi-view learning, our SDL is applicable to a variety of tasks and models, and is superior to alternative decorrelation losses including XCov and DeCov.

\section{Soft CCA}\label{Methodology}

\subsection{Deep CCA} 
\begin{figure}[t]
\centering    
\includegraphics[width=0.6\columnwidth]{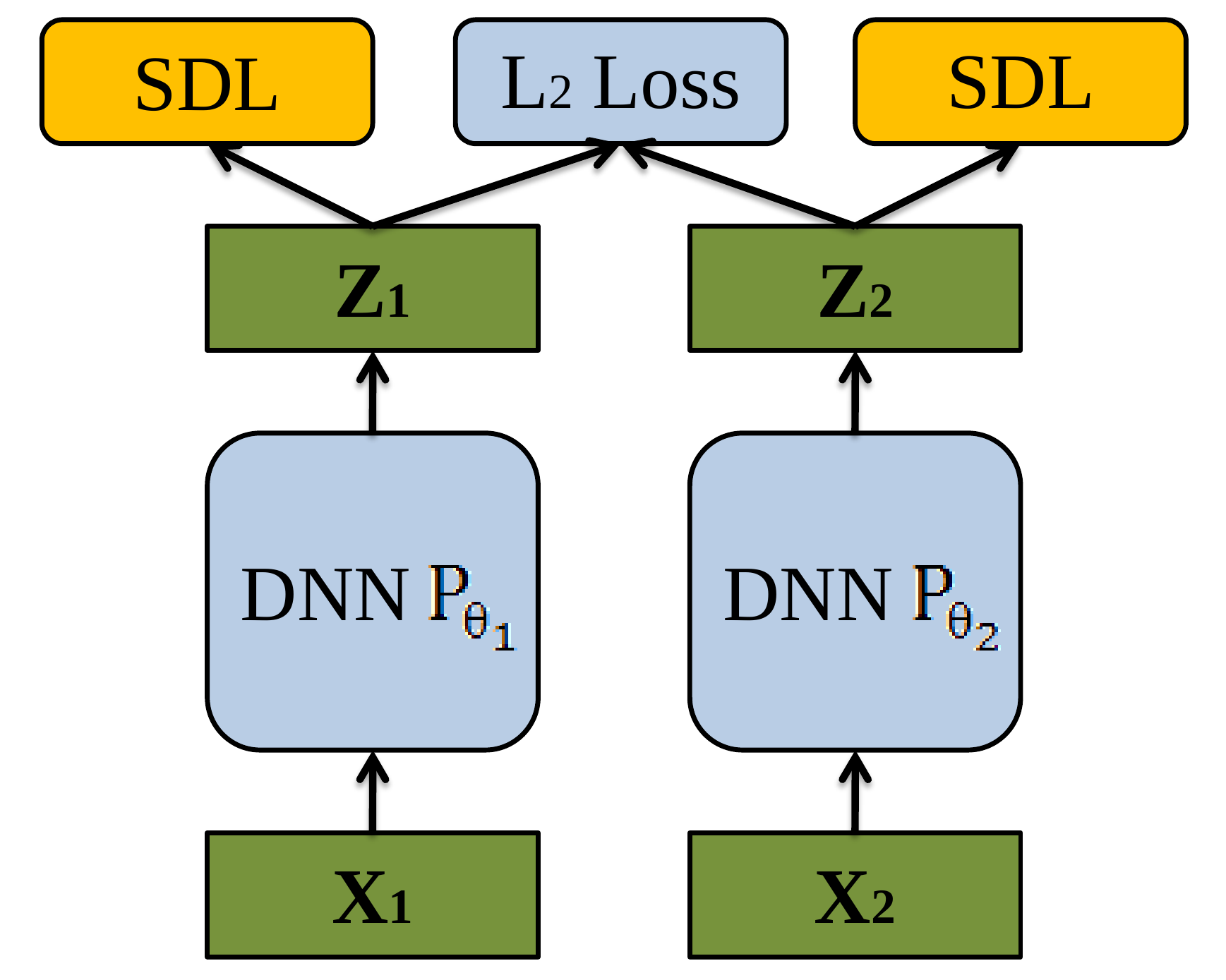}
\caption{Schematic of implementing Soft CCA with SDL.}
\label{DeepCCA_SDL_architecture}
\end{figure}

Deep CCA extends linear CCA model by projecting views of the same item (here we consider images of the same objects) from different views to a common latent space using a DNN with multiple branches, each corresponding to one view (see Fig.~\ref{DeepCCA_SDL_architecture}).

We consider a two-view case for simplicity of notation, but the multi-view extension is straightforward. Assume we have $2N$ images consisting of two views for each of  $N$ objects. They are then organised into mini-batches of $M$ image pairs and fed into the two DNN branches.    The training images in the two views are denoted as $X_1$ and $X_2$ respectively. The DNN branches aim to learn functions that project paired input images into a shared latent space where they are maximally correlated. Denote the DNN projection function for view $i$, $i=\{1,2\}$ as  $P_{\theta_{i}}: X_{i} \to Z_{i}$, or $P_{\theta_{i}}(X_{i}) = Z_{i}$ where $Z_{i} \in \mathbb{R}^{M \times k}$ is the projected feature matrix for $M$ data items for view $i$ in the $k$-D CCA embedding space and $\theta_{i}$ are the DNN parameters.

Following \cite{cca_theory}, CCA can be formulated in multiple ways and the most relevant one here is:
\begin{equation} \label{CCA1}
\begin{split}
\arg \max_{\theta_{1}, \theta_{2}} &\ Tr(P_{\theta_{1}}^{T}(X_1)P_{\theta_{2}}(X_2)),\\
s.t. \ P_{\theta_{1}}^{T}(X_1) & P_{\theta_{1}}(X_1) = P_{\theta_{2}}^{T}(X_2) P_{\theta_{2}}(X_2) = I,
\end{split}
\end{equation}
where $I$ indicates the identity matrix. The constraints enforce decorrelation within each of the two input signals. Eq.~\ref{CCA1} can be written into an equivalent form:
\begin{equation} \label{CCA2}
\begin{split}
\arg \min_{\theta_{1}, \theta_{2}} &\ \frac{1}{2} ||P_{\theta_{1}}(X_1) - P_{\theta_{2}}(X_2)||_{F}^{2}, \\
s.t. \ P_{\theta_{1}}^{T}(X_1) & P_{\theta_{1}}(X_1) = P_{\theta_{2}}^{T}(X_2) P_{\theta_{2}}(X_2) = I, 
\end{split}
\end{equation}
where $||\cdot||_{F}$ is the Frobenius norm of a matrix. It shows that the goal of maximising correlation between $P_{\theta_{1}}(X_1)$ and $P_{\theta_{2}}(X_2)$ can be achieved by minimising the $L_2$ distance between the decorrelated signals.

The key idea of our approach is to convert the hard constraint in Eq.~\ref{CCA2} into a soft cost to be optimised by SGD.

\subsection{Stochastic Decorrelation Loss (SDL)}
%In order to decorrelate the signals, we follow the soft decorrelation paradigm, which formulates the decorrelation procedure as a loss and optimising it along with other losses, rather than the hard one as in \cite{wangSDCCA2015} for computational efficiency.
We denote the representations from one branch of a deep CCA network over a mini-batch as $Z \in \mathbb{R}^{m \times k}$, where $m$ is the mini-batch size and $k$ indicates the number of neurons/feature channels. We further assume that $Z$ has been batch-normalised, i.e., each activation over the mini-batch has zero mean and unit variance. This can be easily achieved by
adding a Batch Normalisation (BN)~\cite{batch_norm} layer. 
%adding a Batch Normalisation (BN)~\cite{batch_norm} layer after the fully connected (FC) layer. 

The mini-batch covariance matrix $C_{mini}^{t}$ for the $t$-th training step then is given as: 
%T: Don't like Cov as a variable name. Normally the variable names are letters, and words represent functions, as in "Compute the covariance"
\begin{equation}\label{cov_mini}
C_{mini}^{t} = \frac{1}{m-1} Z^{T}Z.
\end{equation}

However, full-batch statistics are required by CCA objective for decorrelation. Therefore, we  approximate the full-batch  covariance matrix $C_{full}$ by accumulating statistics collected from each mini-batch. This is achieved by  stochastic incremental learning. More specifically, we first compute an accumulative covariance matrix:
\begin{equation}\label{cov_full_appx}
C_{accu}^{t} = \alpha C_{accu}^{t-1} + C_{mini}^{t},
\end{equation}
where $\alpha \in [0, 1)$ is a forgetting/decay rate and $C_{accu}^{0}$ is initialised with an all-zero matrix. A normalising  factor is also computed accumulatively as $c^{t} = \alpha c^{t-1} + 1$ ($c^{0} = 0$ initially). 
The final full-batch covariance matrix approximation is then computed as:

\begin{equation}\label{cov_final}
C_{appx}^{t} = \frac{C_{accu}^{t}}{c^{t}}.
\end{equation}

If we were to follow an exact decorrelation strategy as in \cite{DCCA_full_batch,DCCA_large_batch,wangSDCCA2015}, we need to force the off-diagonal elements of $C_{appx}^{t}$ to zero. However, that has implications on the computational cost and scalability which we shall detail later. Instead, we follow a soft decorrelation procedure and formulate the decorrelation constraint as a loss. 
%rather than a constraint (requiring an exact decorrelating operation) and encourage soft decorrelation.
Specifically, SDL is an $L_1$ loss on the off-diagonal element of $C_{appx}^{t}$:

\begin{equation}\label{SDL}
L_{SDL} = \sum_{i = 1}^{k}\sum_{j \neq i}^{k} |\phi_{ij}^{t}|,
\end{equation}
where $\phi_{ij}^{t}$ is the element in $C_{appx}^{t}$ at $(i, j)$. $L_1$ loss is used here to encourage  sparsity in the off-diagonal elements.  SDL is soft because it only penalises the correlation across activations instead of enforcing exact decorrelation. It will be jointly optimised with any other losses the model may have.

\noindent\textbf{Gradients and Optimisation}\quad
The gradient of $L_{SDL}$ w.r.t. $z_{ni}$ (the element in $Z$ at $(n, i)$) can be computed as
\begin{equation} \label{SDL_Grad2}
\begin{split}
\frac{\partial L_{SDL}}{\partial z_{ni}} &= \frac{1}{c^{t}} \frac{1}{m-1} \sum_{j}^{k} S(i,j) z_{nj},\\
S(i,j) &= \left\{
                \begin{array}{ll}
                  1,\ \phi_{ij}^{t} > 0\\
                  0,\ i=j \ or \ \phi_{ij}^{t} = 0\\
                  -1,\ \phi_{ij}^{t} < 0
                \end{array}
              \right. \\
\end{split}
\end{equation}
with the sign matrix $S \in \mathbb{R}^{k \times k}$ and $i,j = {1,...,k}$.
Eq.~\ref{SDL_Grad2} can be written  in a matrix form: 

\begin{equation}\label{SDL_Grad3}
\frac{\partial L_{SDL}}{\partial Z} = \frac{1}{c^{t}} \frac{1}{m-1} Z \cdot S,
\end{equation}
where $\cdot$  indicates matrix multiplication.
  
Once the SDL gradients are computed, they are passed through the network during back-propagation and optimised along with other losses in end-to-end training.   

\subsection{Computational Complexity}
%mainly on SDL. Compare with hard decorrelation. Point out which procedure is it about.
Eq.~\ref{SDL} shows that to compute the SDL in a forward pass, we need matrix multiplication (as in Eq.~\ref{cov_mini}), matrix addition (as in Eq.~\ref{cov_full_appx}) and matrix element-wise summation (as in Eq.~\ref{SDL}). Therefore, the forward pass computation complexity of SDL is $O(mk^{2})$.
The gradient computation during the backward pass is in Eq.~\ref{SDL_Grad3}. It is also a matrix multiplication and therefore the complexity is $O(mk^{2})$. The overall computational complexity of one training iteration is thus $O(mk^{2})$. In contrast, existing exact decorrelation computation \cite{DCCA_full_batch,wangSDCCA2015} has a complexity of $O(mk^2 + k^{3})$ due to the use of SVD. Note that in large scale vision problems, the number of activations in an FC layer can easily be thousands, meaning that the alternative hard decorrelation models are prohibitively expensive. 

\subsection{SDL for Soft CCA} 
With the proposed SDL, the constrained optimisation problem in Eq.~\ref{CCA2} can be reformulated as the following unconstrained objective:
\begin{equation} \label{CCA2_eq_CCA1}
\begin{split}
& \arg \min_{\theta_{1}, \theta_{2}} L_{dist}(P_{\theta_{1}}(X_1), P_{\theta_{2}}(X_2)) \\
&  + \lambda ( L_{SDL}(P_{\theta_{1}}(X_1)) + L_{SDL}(P_{\theta_{2}}(X_2))),
\end{split}
\end{equation}
where $L_{dist}(P_{\theta_{1}}(X_1), P_{\theta_{2}}(X_2))$ is the $L_2$ distance  and $\lambda$ weights the alignment versus decorrelation losses. The Soft CCA architecture is also illustrated in Fig.~\ref{DeepCCA_SDL_architecture}. Note that both SDL and $L_2$ loss are mini-batch based losses. Therefore,  Soft CCA (deep CCA model with SDL) can be realised using standard SGD optimisation for end-to-end learning. 

\section{Applications of SDL to other deep models}

\subsection{Factorisation Autoencoder with SDL}
\label{sec:fae}
\begin{figure}[t]
\centering    
\includegraphics[width=0.48\textwidth, height=0.21\textwidth]{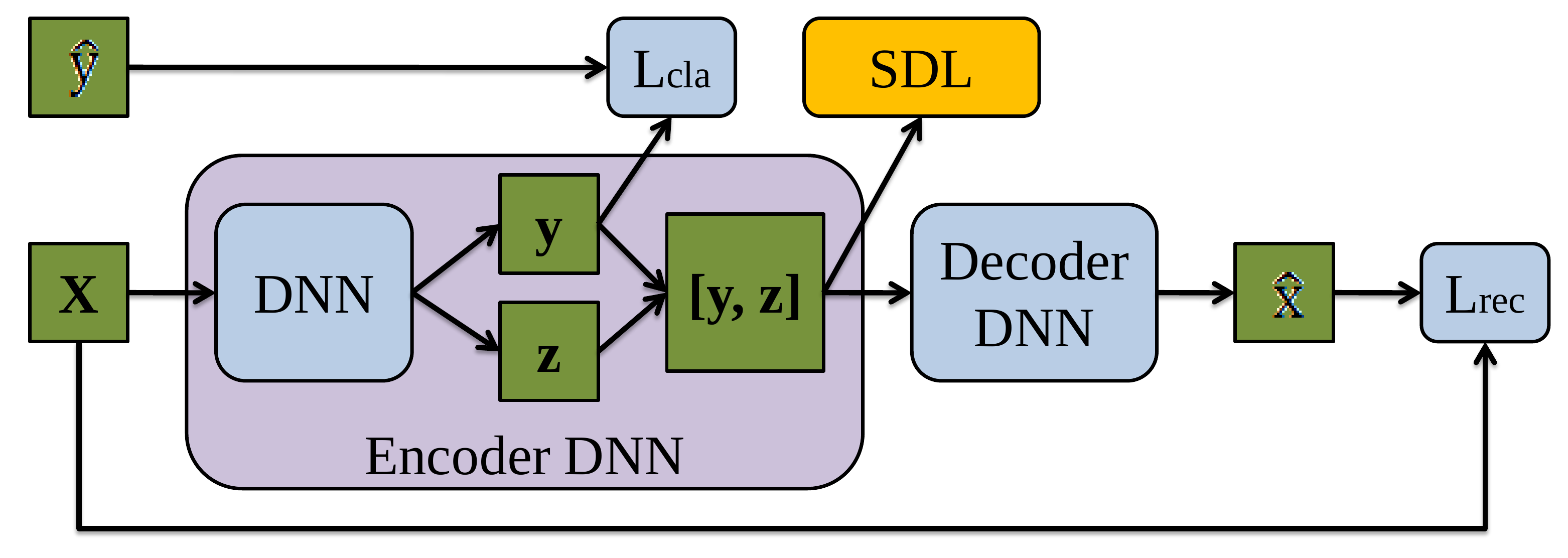}
\caption{Architecture of FAE with SDL.}
\label{Basic_FAE_architecture}
\end{figure}

%Text \& illustration graph.
We describe a two-factor case although the model generalises to an arbitrary number of factors. The two-factor FAE model is illustrated in Fig.~\ref{Basic_FAE_architecture}. Its encoder (a deep neural network) takes image $x$ as input and projects it into an embedding space/latent code which has two parts: $y$ and $z$. We assume $y$ is a factor that is annotated in the training data, e.g., class label. The other unspecified factors are thus captured by $z$. Both $y$ and $z$ are used as input to the decoder (e.g., a deconvolutional network) which produces a reconstruction of $x$, denoted as $\hat{x}$. The goal is not only to accurately reconstruct the input $x$, but also to represent distinct factors of variation in $y$ and $z$ (e.g., class and style respectively).

Assume the FAE model is parameterised by $\theta$. Given a training set $D$ containing images $X$ and their labels $\hat{Y}$ for the known factor, the learning objective of FAE is:
  \begin{equation} \label{FAE1}
\arg \min_{\theta} L_{rec}(X,\hat{X}) 
 + \lambda L_{cla}(Y,\hat{Y}),
\end{equation}
where $ L_{rec}(X,\hat{X})$ is the reconstruction loss, which we use pixel $L_2$ loss here, and $L_{cla}(Y,\hat{Y})$ is the classification loss, i.e., cross-entropy loss here. If there is no constraint on the relation between $y$ and $z$, they would not necessarily represent distinct aspects of the input signal. To disentangle them, we introduce our SDL to the objective:
  \begin{equation} \label{FAE2}
\arg \min_{\theta} L_{rec}(X,\hat{X}) 
 + \lambda_1 L_{cla}(Y,\hat{Y}) +\lambda_2 L_{SDL}([Y,Z]).
\end{equation}
As shown in Fig.~\ref{Basic_FAE_architecture}, this means we decorrelate the elements of the concatenated code $[y,z]$ which decorrelates the two code parts (factors), as well as the signal within the factors.

\subsection{CNN Classifier with SDL}
Since decorrelation loss encourages a layer's activations to be decorrelated, it reduces activation
co-adaptation and maximises the model's capacity. Therefore, SDL can be applied to each layer of a CNN classifier to boost the model performance. In our experiments, we add SDL to different CNN classifiers for different recognition tasks to demonstrate its general applicability.  %Our SDL can be added to different layers to maximise the model capacity and make the model more effective. %<< REdundant.

%\noindent\textcolor{black}{\textbf{DeCov~\cite{decov_l2_loss} and Its Varaiants}\quad SDL has two key differences to existing decorrelation loss DeCov~\cite{decov_l2_loss}:  (i) SDL approximates the global covariance by accumulating mini-batch covariance statistics; and (ii)  SDL exploits L1 instead of L2 loss in DeCov~\cite{decov_l2_loss} for robustness and correlation sparsity. To investigate the impacts of these two differences, two variants of DeCov~\cite{decov_l2_loss}, are proposed. The first loss, called DeCovL1, is proposed by replacing L2 loss with L1 on covariance matrix off-diagonal elements, as in Eq.~\ref{SDL}. The other loss, called DeCovGC, approximates global covariance with mini-batch statistics. Ablation experiments on classification task (in Experiments Section) show that SDL consistently achieves better results than its counterparts due to both of our contributions.}

\section{Experiments}
\label{Experiments}

\subsection{Soft CCA}
\label{sec:exp_cca}

\begin{figure}[t]
  \centering
  \begin{subfigure}[h]{0.8\columnwidth}
    \centering
    \includegraphics[width=0.9\textwidth, height=0.6\textwidth]{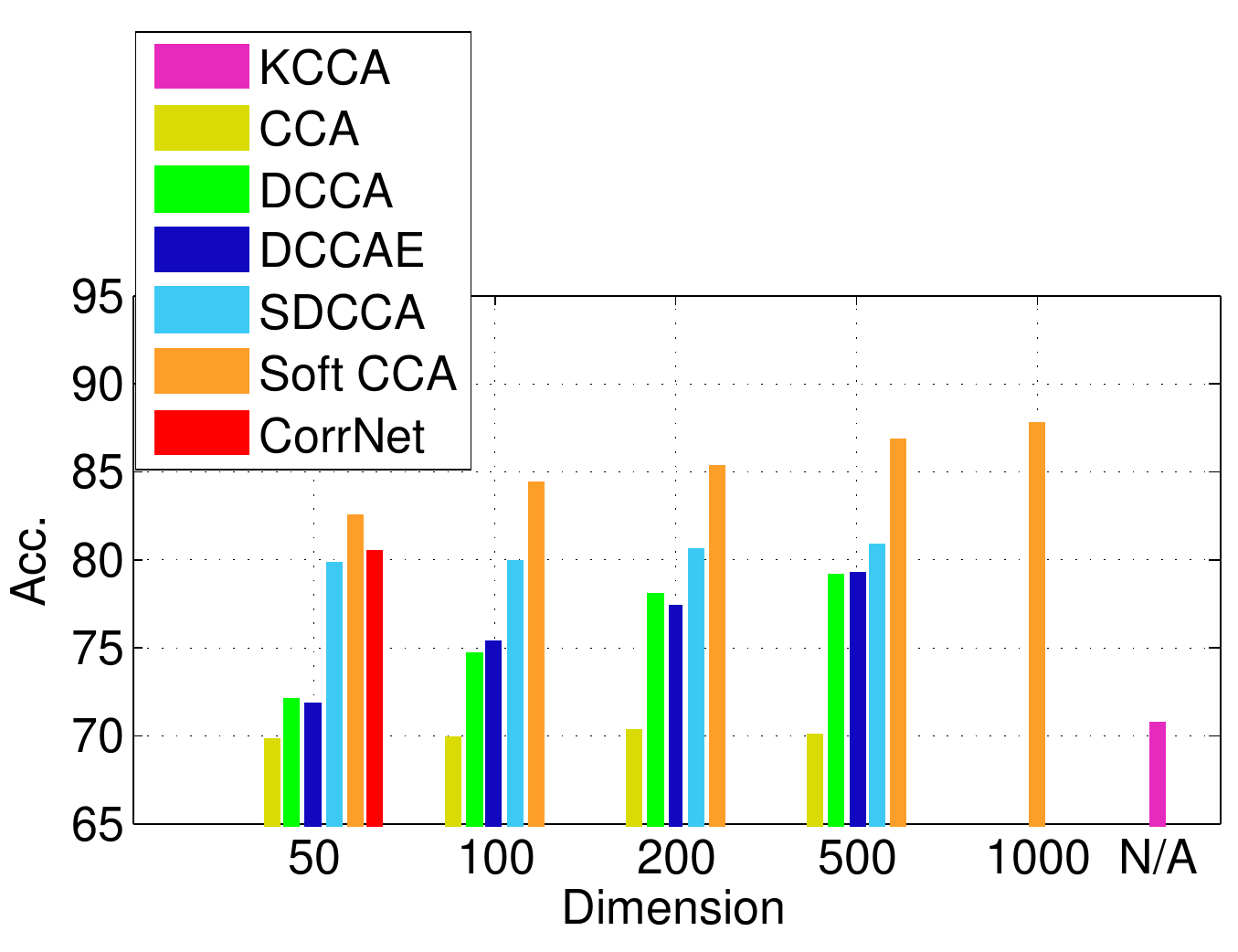}
    \caption{Left-to-right}
    \label{MNIST_transfer_cmp:L2R}
  \end{subfigure}
  \hfill
  \\
  \begin{subfigure}[h]{0.8\columnwidth}
    \centering
    \includegraphics[width=0.9\textwidth, height=0.6\textwidth]{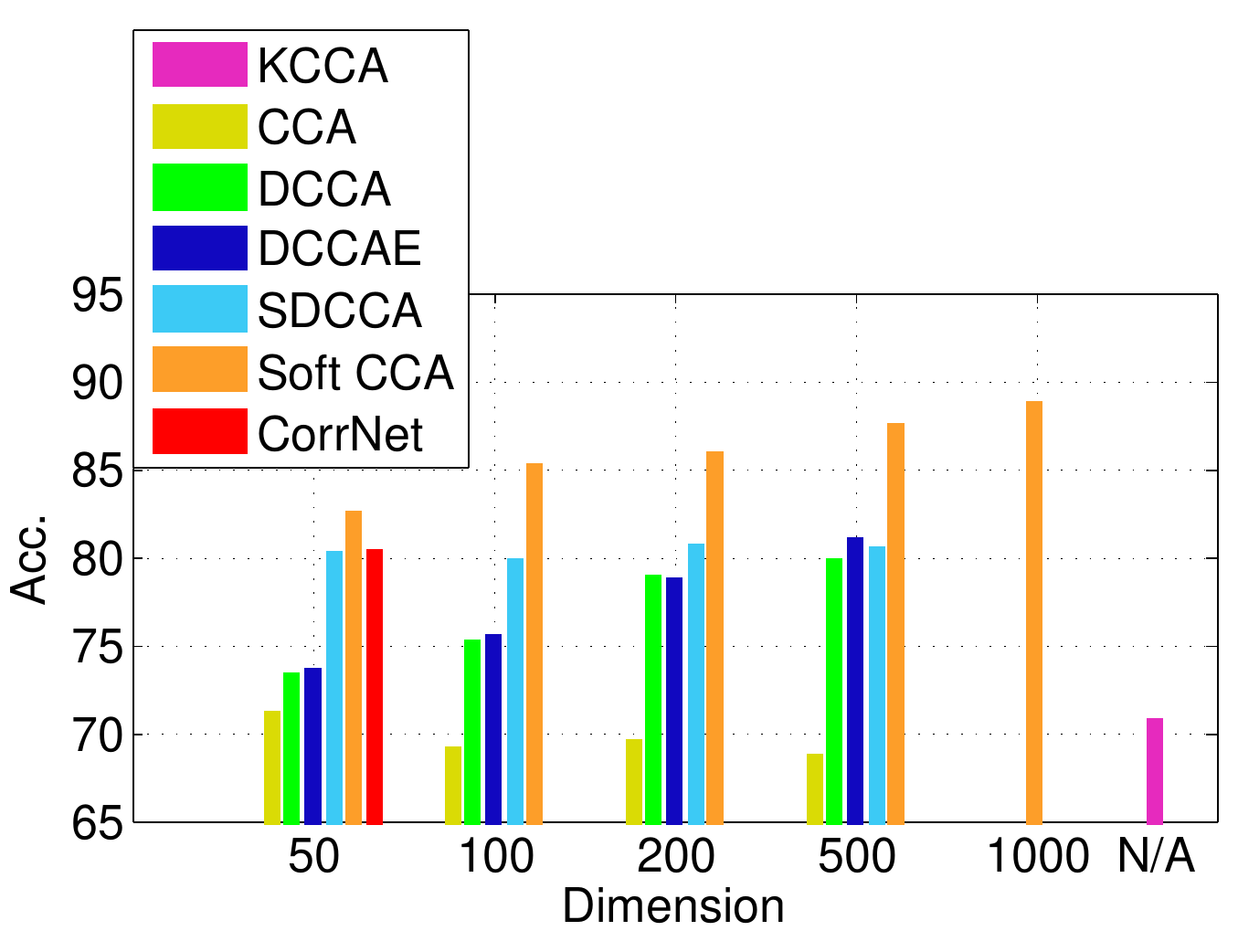}
    \caption{Right-to-left}
    \label{MNIST_transfer_cmp:R2L}
  \end{subfigure}
  \caption{\textcolor{black}{Cross-view digit recognition results on MNIST.  Note that CCA is not scalable to a common space dimension that is greater than the total dimension of 784. Moreover, DCCA, DCCAE and SDCCA are also intractable with our GPU resources when the common space dimension becomes 1000.}}
  \label{MNIST_transfer_cmp}
\end{figure}

\begin{figure}[t]
  \centering
  
  \begin{subfigure}[h]{0.8\columnwidth}
    \centering
    \vspace{0.66cm}
    \includegraphics[width=0.92\textwidth, height=0.5\textwidth]{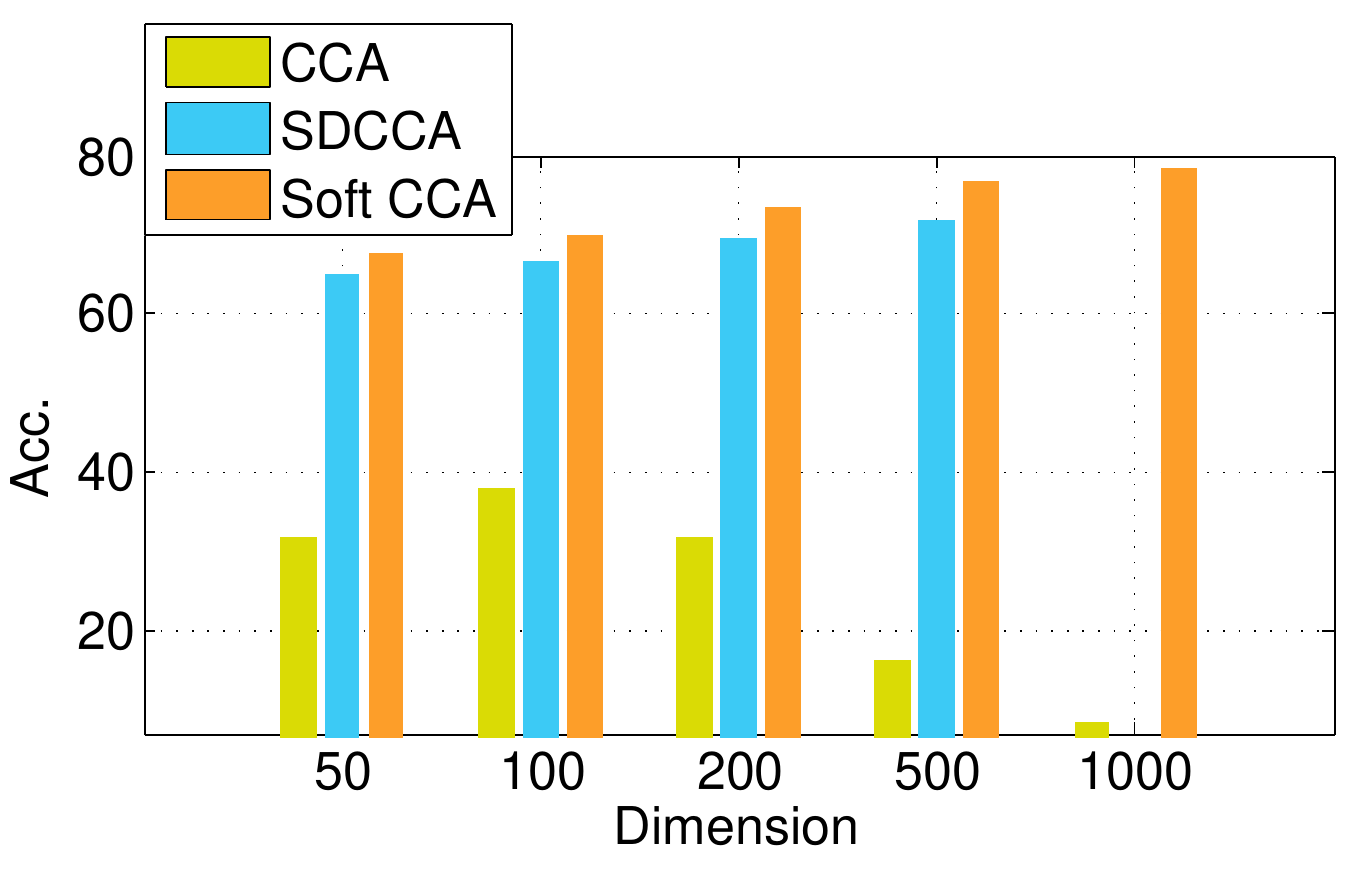}
    \caption{Left-to-right}
    \label{MultiPIE_transfer_cmp:L2R}
  \end{subfigure}
  \hfill
  \begin{subfigure}[h]{0.8\columnwidth}
    \centering
    \vspace{0.66cm}
    \includegraphics[width=0.92\textwidth, height=0.5\textwidth]{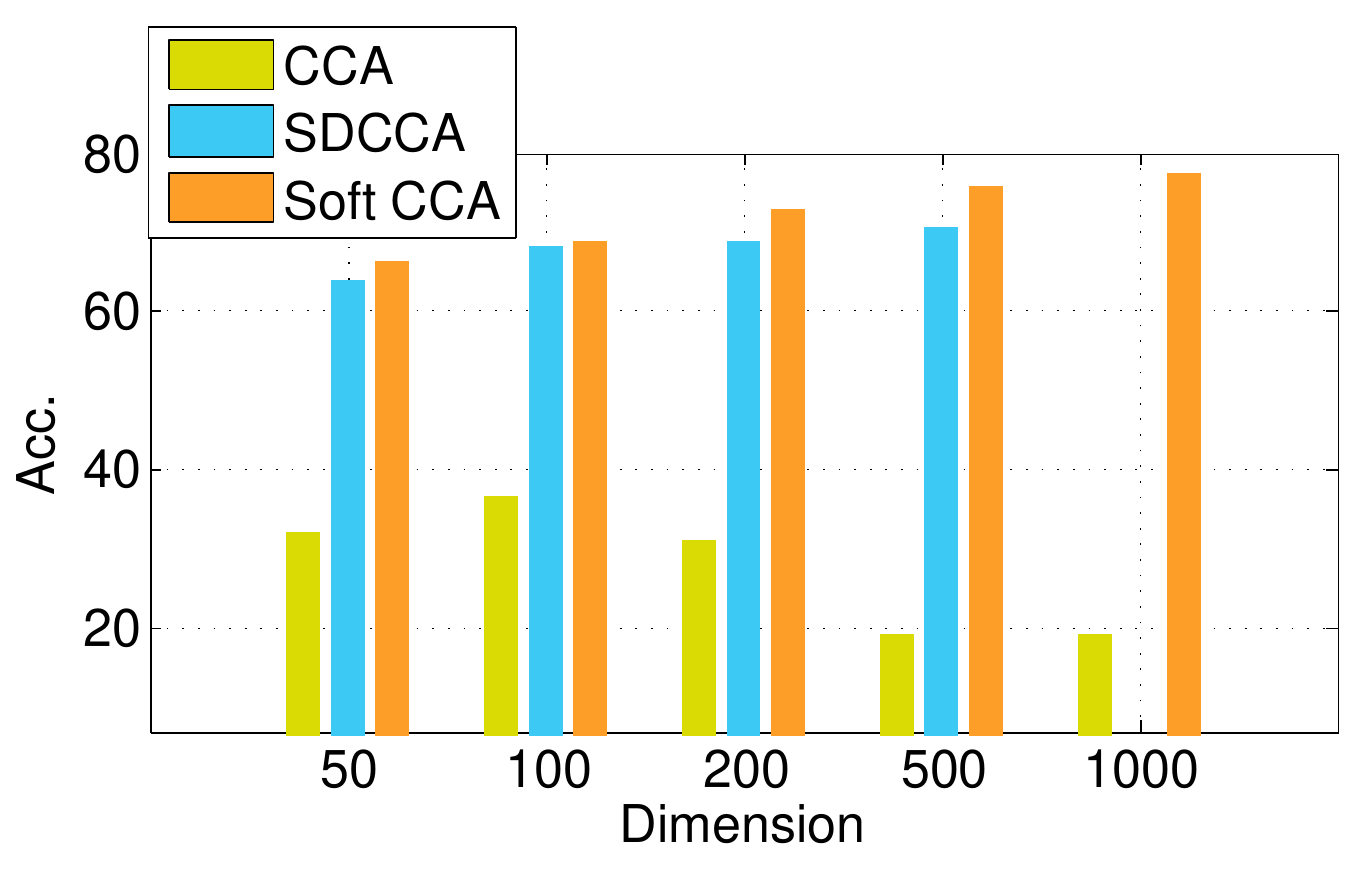}
    \caption{Right-to-left}
    \label{MultiPIE_transfer_cmp:R2L}
  \end{subfigure}
  \caption{Cross-view face recognition results on Multi-PIE. Accuracy (\%). Note that SDCCA is intractable with our GPU resources when the common space dimension becomes 1000. }
  \label{MultiPIE_transfer_cmp}
\end{figure}

%\begin{table}[t]
%%\small
%\centering
%\begin{tabular}{cccc}
%\hline
%                         & Dim 50 & Dim 100        & Dim 200        \\ 
%Upper Bound              & 50     & 100            & 200            \\ \hline
%CCA~\cite{Basic_CCA}                      & 28.3/12.8   & 34.2 /23.9          & 48.7/53.4           \\ 
%%Soft CCA (DeCov)         &  44.2/25.7  & 84.3/56.5 & 162.6/152.3 \\ 
%%Soft CCA (DeCovL1)                     &  44.7/27.8  & 85.7/58.0 & 164.6/153.5 \\ 
%SDCCA~\cite{wangSDCCA2015}                    & {\bf 46.4}/25.7   & {\bf 89.5}          /51.5 & 166.1/151.2          \\ 
%%L2 Loss only             & 8.1    & 18.2           & 39.0           \\ \hline
%%Soft CCA (SDL)           & 45.5/{\bf 29.2}   & 87.0/{\bf 60.5}           & {\bf 166.3}/{\bf 163.2} \\\hline
%Soft CCA           & 45.5/{\bf 29.2}   & 87.0/{\bf 60.5}           & {\bf 166.3}/{\bf 163.2} \\\hline
%\end{tabular}
%\caption{Correlation strength results for MNIST and Multi-PIE $(\uparrow)$ . Expressed as MNIST/Multi-PIE.}
%\label{MNIST_Corr_num}
%\end{table}

\vspace{0.3cm}
\noindent\textbf{Datasets and settings} \quad We evaluate the proposed Soft CCA and alternative deep CCA models on two widely used datasets.  
\textbf{MNIST}~\cite{MNIST_Data} consists of  handwritten digit images with an image size of \textcolor{black}{28 $\times$ 28}. It contains 60,000 training and 10,000 testing images respectively. 
We follow the experimental setting in \cite{CorrNet} for cross-view recognition. Deep CCA models are trained on the left and right halves of a 10,000 sized subset of training images and we do 5-fold cross validation on the provided test set for recognition.
%We follow the cross-view recognition experimental setting in \cite{CorrNet} and train on left and right halves of 10,000 training images and use the provided test set for testing. 
\textbf{Multi-PIE}~\cite{MultiPIE_Data} is a face dataset composed of 750,000 images of 337 people with various factors contributing to appearance variation including viewpoint, illumination and facial expression. We use a subset containing 6,200 images of all 337 identities in neutral expression and lighting. Constructing an analogous experiment to the cross-view recognition benchmark, these images are separated into the left and right view groups according to their viewing angle. Left-right view angle pairs are then formed exhaustively for the same identities to train the deep CCA models.  
%we use half of the images in both views for model training and the rest for testing.
We use half of the images in both views for deep CCA training and also do 5-fold cross validation for recognition on the rest of the data.

\vspace{0.3cm}
\noindent\textbf{Implementation details} \quad For MNIST cross-view recognition, the network architecture of each view branch is identical to that in \cite{CorrNet} for fair comparison. Concretely, there are three hidden layer containing 500, 300, $k$ units/activations respectively, where the $k$ units are used as the common representation (CCA embedding layer). ReLU is applied on the hidden layers' activations (except the embedding layer). %In our experiment, $k$ is set to 50, 100 and 200.
Once the CCA model is trained, on the test set, features from one view (e.g., right) are exacted, embedded with deep CCA, and then fed to a Linear SVM~\cite{libsvm} classifier which is trained to recognise the images. Finally, the model is evaluated based on features from the other view (e.g.,~left) being projected into the shared embedding space, and recognised by the SVM. Clearly, the performance of the SVM on this cross-view recognition task depends on the efficacy of the CCA embedding. 
\textcolor{black}{An analogous cross-view recognition setting is used for the Multi-PIE dataset. The DNN architecture for Multi-PIE also has three hidden layers: 1024, 512, $k$ units, the $k$ units are used as the CCA embedding layer. ReLU is applied on the hidden layers' activations (except the embedding layer).} %The cross-view recognition procedure is similar to that on MNIST.}

\vspace{0.3cm}
\noindent\textbf{Competitors} \quad For shallow CCA, we compare the standard linear CCA~\cite{Basic_CCA} and its nonlinear kernelised variant, KCCA~\cite{KCCA}.
\textcolor{black}{The KCCA results are obtained from \cite{CorrNet}.}
%KCCA is not tractable with the computer hardware we have when $k>50$, so only results with $k=50$ are reported.
For the deep CCA models, \textcolor{black}{we compare with CorrNet~\cite{CorrNet}, DCCA~\cite{DCCA_full_batch}, DCCAE~\cite{wang2015-deep-multi-view} and SDCCA~\cite{wangSDCCA2015}.}  CorrNet \cite{CorrNet} combines correlation maximisation with cross-view autoencoder loss and uses Batch Normalisation. 
 %but for the decorrelation, 
Without access to their code, we can only use the reported result in \cite{CorrNet} which was obtained only on MNIST with $k=50$. As far as we know, SDCCA~\cite{wangSDCCA2015} is the most efficient state-of-the-art deep CCA model to date.  
%\textcolor{black}{Moreover, we also replace SDL with DeCov~\cite{decov_l2_loss} loss or  DeCovL1 loss for comparison.}

\begin{table}[t]
\small
\centering
\begin{tabular}{cccccc}
\hline
                         & 50D & 100D        & 200D   & 500D & 1000D     \\  \hline
Upper Bound              & 50     & 100            & 200        &  500    & 1000  \\ \hline
CCA~\cite{Basic_CCA}             & 28.3   & 34.2          & 48.7    &  74.0   &   -\\ 
DCCA~\cite{DCCA_full_batch}        & 29.5   &  44.9  &   59.0  &  84.7 & - \\
DCCAE~\cite{wang2015-deep-multi-view}       & 29.3   &  44.2  &   58.1  &  84.4 & - \\
SDCCA~\cite{wangSDCCA2015}       & {\bf 46.4}   & {\bf 89.5}  & 166.1    & 307.4 &   -   \\ 
Soft CCA                         & 45.5   & 87.0 & {\bf 166.3} & {\bf 356.8} & {\bf 437.7} \\ \hline
\end{tabular}
\caption{\textcolor{black}{Correlation strength on MNIST. `-' indicates that the result is not obtainable due to the corresponding model being intractible with our available hardware.}}
\label{tab:MNIST_CorrScore}
\end{table}

\vspace{0.3cm}
\noindent\textbf{Results on cross-view recognition} \quad Figures~\ref{MNIST_transfer_cmp} and \ref{MultiPIE_transfer_cmp} show the results for cross-view digit and face recognition. We make the following observations: 
(1) The deep models achieve better performance than the shallow ones.
(2) Our Soft CCA achieves the best results on both datasets with all CCA space dimensions.
(3) \textcolor{black}{Increasing the common space dimension $k$ benefits SDCCA very little and even harms the performance of other competitors (e.g. CCA). In contrast,
our Soft CCA clearly benefits from larger CCA space dimensions.
} %Fig.~\ref{MultiPIE_transfer_cmp} further illustrates the effectiveness of Soft CCA under large dimensions, e.g. 500 and 1000.}
%Both CCA and SDCCA benefit little from increasing  CCA space dimension, whilst our model clearly benefits.
%\textcolor{black}{4) Soft CCA with SDL achieves the best performance overall among competitors on cross view recognition. 
%and correlation strength
%5) SDL outperforms DeCovL1, which in turn outperforms vanilla DeCov~\cite{decov_l2_loss}. This demonstrates that both our contributions to L1 penalty and incremental correlation statistics are beneficial.}

\vspace{0.3cm}
\noindent\textbf{Results on cross-view correlation} \quad 
Another way to evaluate CCA models is to measure the average correlation strength of each matching pair of data when they are projected into the common CCA space \cite{wangSDCCA2015}. 
We follow the experimental setting and network architecture of \cite{wangSDCCA2015} (SDCCA) for a fair comparison.
%. Experiments on Multi-PIE follows the similar setting of cross-view recognition. (xb: do we need such details? It seems messy and distracted if put it in the former part.)}
%The results on both datasets are shown in Table~\ref{MNIST_Corr_num}. 
\textcolor{black}{The results of MNIST and Multi-PIE are shown in Table~\ref{tab:MNIST_CorrScore} and Table~\ref{tab:MultiPIE_CorrScore} respectively.}
We can conclude from the results that: (1) Again the deep models achieve higher correlation values indicating that they align the two views much better than the linear CCA model. (2) For the easier digit classification task in MNIST, our model is slightly inferior to SDCCA at 50D and 100D, but better after 200D. For the more challenging face recognition problem in Multi-PIE, Soft CCA consistently outperforms SDCCA and the gap increases with the dimension. These results suggest that our model is more effective with higher dimensional embedding space, which is required for more challenging computer vision tasks.   

\begin{table}[t]
\small
\centering
\begin{tabular}{cccccc}
\hline
                         & 50D &  100D        & 200D   & 500D & 1000D     \\  \hline
Upper Bound              & 50     & 100            & 200        &  500    & 1000  \\ \hline
CCA~\cite{Basic_CCA}                      & 12.8   & 23.9          & 53.4    &  140.6   &   207.1    \\ 
SDCCA~\cite{wangSDCCA2015}                    & 25.7   & 51.5 & 151.2    & 228.3 &   -   \\ 
Soft CCA           & {\bf 29.2}   & {\bf 60.5}           & {\bf 163.2} & {\bf 257.7} & {\bf 283.9} \\ \hline
\end{tabular}
\caption{\textcolor{black}{Correlation strength on Multi-PIE. }
}
\label{tab:MultiPIE_CorrScore}
\end{table}

\begin{figure}[t]
\centering    
\includegraphics[width=0.9\columnwidth]{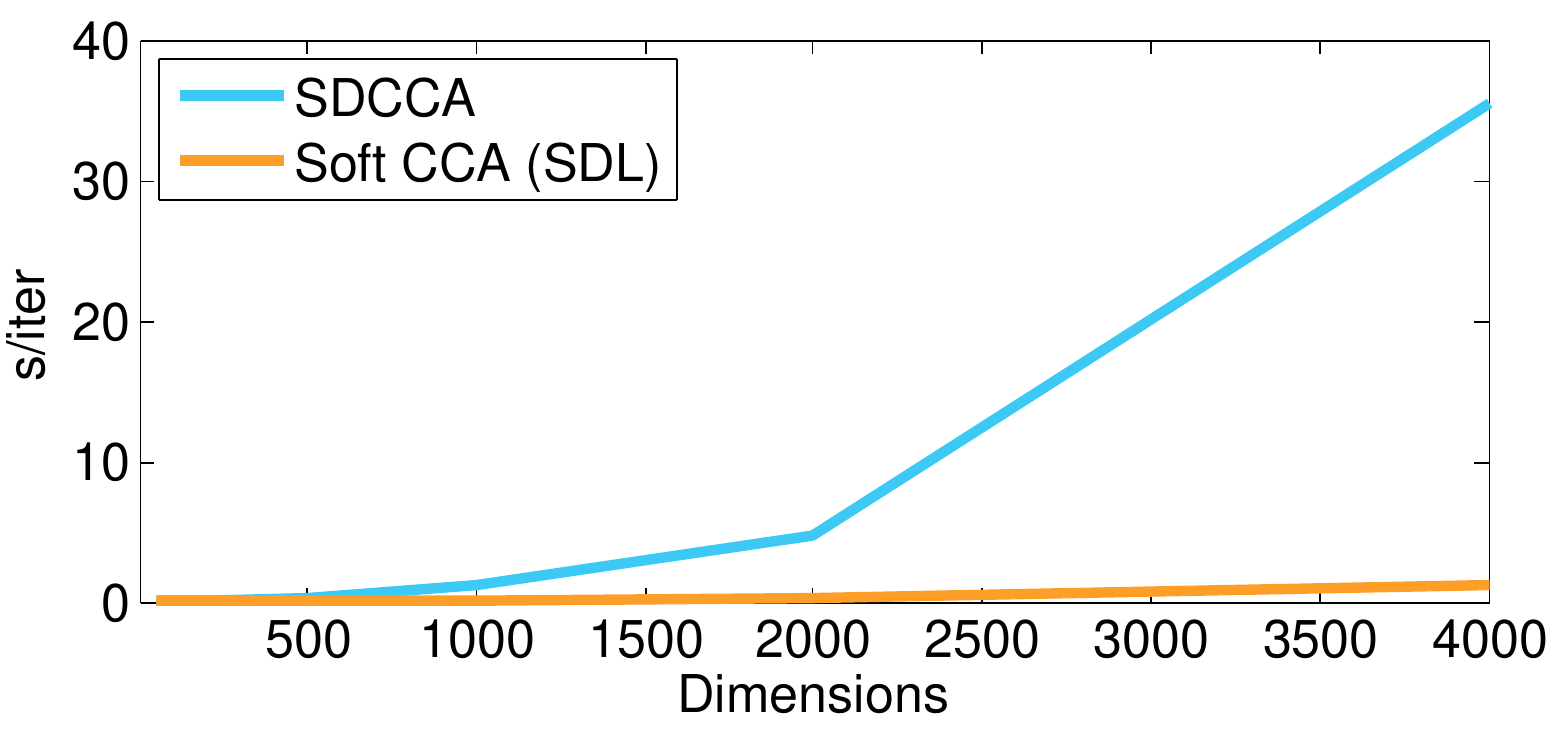}
\caption{Comparing training time (seconds/iteration) on MNIST given different CCA space dimensions.}
\label{MNIST_Train_Time_cmp}
\end{figure}
\vspace{0.3cm}
\noindent\textbf{Evaluation on scalability} \quad 
We compare the training time for our model and that for the most efficient deep CCA model proposed to date, SDCCA \cite{wangSDCCA2015}.  Figure \ref{MNIST_Train_Time_cmp} shows that  our soft CCA is always more efficient than SDCCA even at the low dimensions\footnote{The speedup is significant even under low dimensions; it is just not very salient in Fig.~\ref{MNIST_Train_Time_cmp} due to the scaling problem. E.g, at 50D and 100D, Soft CCA is 2 and 5 time faster to train respectively. }. Importantly,  when the CCA embedding space dimension approaches 4,000 (roughly the same as the final FC layer size of popular DNNs like AlexNet and VGGNet), our model is clearly much more efficient to train. This is due to the $O(k^2)$ vs.~$O(k^3)$ computational complexity difference.

\subsection{FAE with SDL}\label{FAE_SDL}

\begin{figure*}[t]
\centering    
\includegraphics[width=1.0\textwidth, height=0.3\textwidth]{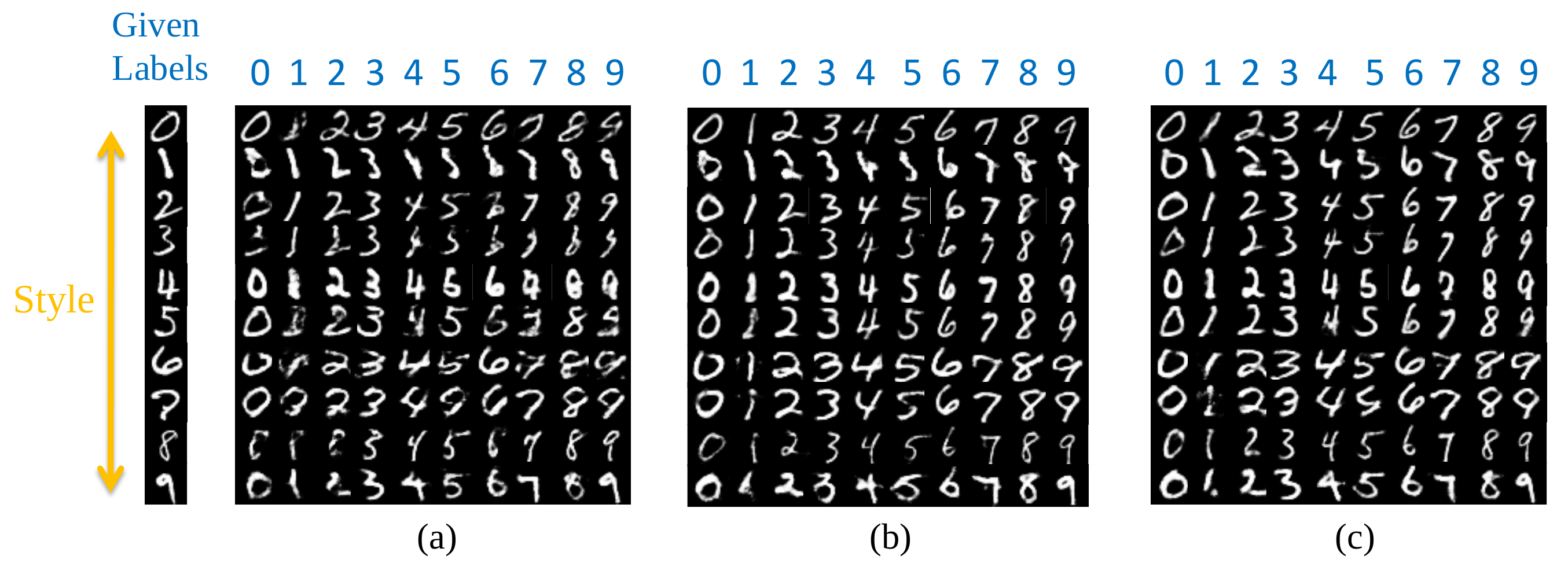}
\caption{Qualitative results of handwriting style transfer with different FAE models. (a) FAE; (b) FAE + XCov~\cite{Discover_Hidden_Factor} ; (c) FAE + SDL. The dimension of z is set to 10.}
\label{FAE_style_illustrate}
\end{figure*} 
\vspace{0.3cm}
\noindent\textbf{Dataset and settings} \quad We use \textcolor{black}{MNIST \cite{MNIST_Data}, and follow the same experimental setting as~\cite{Discover_Hidden_Factor}. 
The network architecture is 784-1000-1000-\{y+z\}-1000-1000-784, where 784 is the dimension of the vectorised image. ReLU is applied on the hidden layers' activations (except y, z)}.
%TH: Reconstruction output is not hidden.
 As shown in Fig.~\ref{Basic_FAE_architecture}, among the two factors to be disentangled, $y$ is the digit class which is annotated with the training data. The other factor $z$ corresponds to aspects of appearance besides class -- i.e., the unannotated writing style. In our experiments, the dimension of $y$  is fixed to 10 corresponding to the 10 digit classes and the dimension of $z$ is also set to 10.
We compare the performance of a vanilla FAE (basic network with only reconstruction and classification loss), FAE+XCov ~\cite{Discover_Hidden_Factor}, FAE+DeCov~\cite{decov_l2_loss} and our FAE+SDL.

 \vspace{0.3cm}
\noindent\textbf{Evaluation on disentanglement} \quad In the ideal case, the two factors will be completed disentangled in $y$ and $z$, i.e., $y$ contains no information about the style and $z$ contains nothing about the class. To quantify this, \textcolor{black}{we compare the digit classification performance with the inferred $y$ and $z$ on the test set. Classification based on $y$ is given by the prediction scores from the FAE classification branch. The inferred $z$ requires an additional classification model and we train a linear SVM using $z$ from the training set and test it on the test set.} 
%we use the inferred $y$ and $z$ to train  SVM classifiers and test them on the test data for digit classification. 
Predictions based on $y$ and $z$ should thus ideally give \textcolor{black}{perfect and random chance accuracies respectively}.  
Table~\ref{Fact_AE_Class_Branch} shows that with SDL, the style feature $z$'s classification performance is close to random guess (10\%), and better (closer to random) than that of XCov and DeCov, whilst using with the vanilla FAE with no decorrelation loss, it still contains extensive class information. Meanwhile, the disentangled $y$ provides the highest classification accuracy using our FAE+SDL. The results suggest that our model is more effective than the alternative XCov and DeCov in disentangling latent factors. This is because our SDL does a stochastic approximation of the full-batch statistics, whilst both XCov and DeCov only use information from each mini-batch.

\begin{table}[t]
%\small
\centering
\begin{tabular}{ccccc}
\hline
                                                         & FAE & XCov~\cite{Discover_Hidden_Factor} & DeCov~\cite{decov_l2_loss} & SDL\\ \hline
$z$ $(\downarrow)$ & 43.44                                                   & 14.51                                                         & 15.42 & {\bf 11.35}                                                          \\ 
$y$  $(\uparrow)$ & 97.23                                                   & 95.72                                                         & 97.09 & {\bf 97.33}                                                          \\ \hline
\end{tabular}
\caption{Disentanglement efficacy. Classification accuracy $(\%)$ using representation of each branch in MNIST FAE. }
\label{Fact_AE_Class_Branch}
\end{table}

\vspace{0.3cm}
\noindent\textbf{Qualitative results} \quad With the style factor disentangled from the class factor, we can use the FAE to transfer styles to a new digit. Given an input image containing a certain digit with certain handwriting style, we can keep the inferred $z$ and change the value $y$ manually to a different digit class. After feeding both the original $z$ and the modified $y$ to the decoder, we can synthesise a new digit with the same style as the input image. Qualitative results are shown in Fig.~\ref{FAE_style_illustrate}. We see the better disentanglement efficacy of our model in terms of clearer digit reconstruction with clearer style transfer.
%\textcolor{black}{In this figure, compare with FAE+XCov if possible.}

\subsection{CNN Classifier with SDL}
\noindent\textbf{Experiments on object recognition} \quad
We use CIFAR10~\cite{CIFAR10_Data} which consists of 60,000 $32\times32$ colour images in 10 categories, with 6000 images per category. We follow the standard experimental setting in \cite{CIFAR10_Data}.  
%The DNN baseline model used is a 10-layer simplification of the 20-layer plain ResNet in \cite{resnet}\footnote{The original network architecture is twice as deep and we remove half of the layers to fit into a single GPU card (NVIDIA 1080P).}.  
\textcolor{black}{The DNN baseline model used is a 20-layer ResNet~\cite{resnet}.}
%We compare the baseline model (Batch Normalisation (BN)~\cite{batch_norm}) with our SDL model. All models use BN on all layers.
%Different decorrelation losses are applied to the baseline model (Batch Normalisation (BN)~\cite{batch_norm}) for comparisons.
%SDL is applied on the activations of each BN layer during training.
%For a fair comparison, we fix the initial states of both models to be the same. 
%and report the averaged performance over 5 trials. 
We compares SDL with existing decorrelation loss DeCov~\cite{decov_l2_loss} and the baseline (with BN but without any decorrelation loss) in Table~\ref{cifar10_classify}.
\textcolor{black}{The proposed SDL leads to a 1.32\% performance improvement over the baseline model and also outperforms the alternative DeCov loss by 0.82\%.} 
%The proposed SDL provides a 1.42\% performance improvement over the baseline model and also outperforms the alternative DeCov loss by 0.90\%. 

%\begin{table}[t]
%%\small
%\centering
%\begin{tabular}{cc}
%\hline
%                    & Accuracy       \\ \hline
%Baseline~\cite{resnet}                  & 84.57          \\ %\hline
%DeCov~\cite{decov_l2_loss}               & 85.09          \\ %\hline
%%DeCovGC             & 85.36          \\ %\hline
%%DeCovL1             & 85.43          \\ %\hline
%SDL                 & {\bf 85.99}          \\ \hline
%\end{tabular}
%\caption{CIFAR10 classification results (\%)}
%\label{cifar10_classify}
%\end{table}

\begin{table}[t]
%\small
\centering
\begin{tabular}{cc}
\hline
                    & Accuracy       \\ \hline
Baseline~\cite{resnet}                  & 91.12          \\ %\hline
DeCov~\cite{decov_l2_loss}               & 91.62          \\ %\hline
%DeCovGC             & 85.36          \\ %\hline
%DeCovL1             & 85.43          \\ %\hline
SDL                 & {\bf 92.44}          \\ \hline
\end{tabular}
\caption{\textcolor{black}{CIFAR10 classification results (\%)}}
\label{cifar10_classify}
\end{table}

\begin{table}[t]
%\small
\centering
\begin{tabular}{ccccc}
\hline
\multirow{2}{*}{}     & \multicolumn{2}{c}{S-Query}    & \multicolumn{2}{c}{M-Query}    \\% \cline{2-5} 
  Method                    & mAP            & R1             & mAP            & R1             \\ \hline
%\cite{market_0}        & 14.75          & 35.84          & 19.42          & 44.36          \\ \hline
%\cite{market_1}            & \textbackslash & 45.16          & \textbackslash & \textbackslash \\ \hline
%\cite{market_2}            & 22.31          & 47.92          & \textbackslash & \textbackslash \\ \hline
%DADM~\cite{market_5}          & 19.6           & 39.4           & 25.8           & 49.0           \\ 
%\cite{market_3}        & 29.87          & 55.43          & 46.03          & 71.56          \\ \hline
%\cite{market_4}     & 26.35          & 51.90          & \textbackslash & \textbackslash \\ \hline
%MSTC~\cite{market_6}      & -- & 45.1           & -- & 55.4           \\ 
LDEHL~\cite{market_7}       & -- & 59.47          & -- & -- \\
Siamese LSTM~\cite{market_8}           & -- & -- & 35.3           & 61.6           \\ 
Gated S-CNN~\cite{market_9}         & 39.55          & 65.88          & 48.45          & 76.04          \\ 
%\cite{market_10}     & 47.89          & 73.87          & 56.98          & 81.29          \\ \hline
%\cite{market_11} & 56.23          & 78.06          & 68.52          & 85.12          \\ \hline
CNN Embedding~\cite{market_12}   & 59.87          & 79.51          & 70.33          & 85.84          \\ 
%Deep Transfer~\cite{market_13}    & 65.5           & 83.7           & 73.8           & 89.6           \\
Spindle~\cite{spindlenet_2017}  &  -      &   76.9     &     -     &      -      \\ %\hline
HP-net~\cite{hydraplus}  &  -       &    76.9   &     -     &      -      \\ %\hline
OIM~\cite{xiao2017joint} &  -       &   82.1   &     -     &      -      \\ %\hline
Re-rank~\cite{rerank_reid} &   63.6    &    77.1 &     -     &      -      \\ %\hline
DPA~\cite{zhao2017deeply}     &    63.4      &  81.0       &   -       &   -         \\ %\hline
SVDNet~\cite{sun2017svdnet}     &   62.1          &   82.3   &     -     &     -       \\ %\hline
%DaF~\cite{DaF}                &     82.3      &   72.4     &     -     &     -       \\ %\hline
ACRN~\cite{reid_attrib_1}       &      62.6     &   83.6     &     -     &     -       \\ %\hline
%Transfer~\cite{dual_loss}     &     83.7      &   65.5     &   89.6       &   73.8         \\ %\hline
Context~\cite{body_parts_reid} &    57.5       &   80.3     &     66.7     &  86.8          \\ %\hline
JLML~\cite{jlml_wei_2017}     &    64.4    &    83.9       &   74.5     &     89.7         \\ %\hline
LSRO~\cite{Duke_reid_dataset} &      66.1     &    84.0      &   \textbf{76.1 }     &     88.4         \\
 \hline 
DGDNet$^{*}$                & 64.55          & 85.06          & 73.30          & 89.40          \\
DGDNet+DeCov~\cite{decov_l2_loss}    & 65.74     & 85.86      & 74.72     &  90.53      \\
%DGDNet+DeCovL1    &    66.57     &    86.01   &   74.72    &   90.53         \\
DGDNet+SDL            & \textbf{67.67}          & \textbf{86.75}          & 75.77        & \textbf{91.06}          \\ \hline
\end{tabular}
\caption{Market-1501 Results. S-Query means Single Query, and  M-Query means Multiple Query. `--' indicates no reported result. DGDNet$^{*}$ refers to the basic network used in DGD~\cite{DGD}, but trained from scratch only on Market-1501, without multi-task learning through the Domain Guided Dropout layer using six auxiliary datasets for fair comparison with the state-of-the-art.}
\label{Market_classify}
\end{table}

\begin{table}[t]
%\small
\centering
\begin{tabular}{ccc}
\hline
                    & CIFAR 10    & Market-1501  \\ \hline
DeCov~\cite{decov_l2_loss}               & 91.62     & 85.86     \\ %\hline
DeCovGC             & 91.86      &    86.28     \\ %\hline
DeCovL1             & 91.90   &    86.01        \\ %\hline
SDL                 & {\bf 92.44}        & \textbf{86.75}    \\ \hline
\end{tabular}
\caption{\textcolor{black}{Ablation study on the advantage of SDL over DeCov. The CIFAR10 classification results are in classification accuracy (\%) and the Market-1501 results  are in R1 accuracy  (\%) under the single query setting.} }
\label{cifar10_classify_ablation}
\end{table}

\vspace{0.3cm}
\noindent\textbf{Person re-identification} \quad
%\noindent\textbf{Datasets and Settings} \quad
In this experiment, a CNN classifier is applied to solve a more challenging recognition problem. 
%In this experiment, we evaluate our SDL on a much more challenging instance recognition problem. 
The person re-identification (Re-ID) problem aims to match pedestrians captured by non-overlapping CCTV cameras\footnote{Note that although Re-ID can be interpreted as a multi-view learning problem, state-of-the-art approaches treat it as an identity-supervised single-view identity classification problem. \cite{DGD};  we thus follow this single-view approach.}. We use one of the biggest and most popular Re-ID benchmarks.  \textbf{Market-1501}~\cite{market_dataset} 
is collected from 6 different cameras.
It has 32,668 bounding boxes of 1,501 identities obtained using a
Deformable Part Model (DPM) person detector. Following the standards
split \cite{market_dataset}, we use 751 identities with 12,936 images
for training and the rest 750 identities with 19,732 images for testing.  Experiments are conducted under both the single-query and
multi-query evaluation settings. The Rank-$1$ accuracy is computed to evaluate all the methods. We also calculate the mean average precision (mAP)~\cite{market_dataset}. For the base model, we use one of the state-of-the-art deep Re-ID models, DGDNet \cite{DGD}, which is built on Inception modules~\cite{google_inception}.
%. DGDNet is a simplified GoogLeNet recently proposed in \cite{DGD}. \textcolor{black}{(DGDNet is based on Inception modules~\cite{google_inception} and recently proposed in \cite{DGD}. xb: this is how DGD paper claims their model.)} 
\textcolor{black}{Our model (DGDNet+SDL) adds SDL on the output of each BN layer in DGDNet during training.}

The results are shown in Table \ref{Market_classify}, along with some recent high performing state-of-the-art alternatives.
We can see that: (1) Our model (DGDNet+SDL)  outperforms a number of state-of-the-art alternatives. (2) Compared to the base model (DGDNet without decorrelation loss), adding our  SDL boosts the performance by a clear margin. 
\textcolor{black}{(3) When the alternative DeCov loss is added to the base model, its performance is also improved, but by a smaller margin. This result thus indicates that the proposed SDL is more effective than DeCov. }

\vspace{0.3cm}
\noindent\textbf{Ablation study} \quad
Note that SDL differs from DeCov in two aspects: (i) SDL approximates the global covariance by accumulating mini-batch covariance statistics; and (ii)  SDL exploits an $L_1$ instead of $L_2$ formulation as in DeCov for robustness and correlation sparsity. In order to gain some insight on what contribute to SDL's  superior performance, we consider two variants of DeCov~\cite{decov_l2_loss}, called DeCovGC and DeCovL1.  DeCovGC is DeCov with added accumulating covariance statistic only while DeCovL1 adopts a $L1$ formulation as in SDL.
As shown in Table~\ref{cifar10_classify_ablation}, both DeCov variants have better results than DeCov~\cite{decov_l2_loss} while SDL (with both accumulating covariance statistic and $L1$ loss) achieves the highest performance among them. It suggests that both differences contribute to the effectiveness of SDL. 

\section{Conclusions}\label{Conclusions}
We have proposed a novel deep CCA model, termed Soft CCA, which provides an efficient and effective solution to deep CCA optimisation by introducing a soft decorrelation loss. Extensive experiments show that the proposed Soft CCA is more effective and scalable than existing CCA variants. Compared to exact whitening solutions, Soft CCA is easy to implement in contemporary learning frameworks, and therefore is promising for enabling practical use of CCA techniques in the deep learning community.  Moreover, we demonstrated that as a by-product, the developed SDL loss can be applied beyond CCA as a general purpose decorrelation loss -- to any deep learning task where feature decorrelation is required. As case studies, SDL was shown to outperform alternative decorrelation losses in \textcolor{black}{FAE latent factor disentanglement and CNN object and instance recognition.}
%both FAE disentanglement and CNN image recognition.

%We propose a novel Deep CCA model, Soft CCA, which provides both efficient and effective solution to deep CCA optimisations based on the Stochastic Decorrelation Loss (SDL). 
%Moreover, SDL can further improve the performance of Factorisation Autoencoder (FAE), a popular latent multi-view learning model (lMVL). 
%Therefore, both eMVL (i.e. CCA) and lMVL (i.e. FAE) can be handled under a unified framework using decorrelation losses (e.g. SDL). The intrinsic connections between decorrelation and multi-view learning (MVL) are revealed. 
%Moreover, SDL's efficacy is demonstrated by the fact that it consistently achieves better performance than existing decorrelation losses on all MVL tasks.

{\small
\bibliographystyle{ieee}
\bibliography{egbib,reid_egbib}
}

\end{document}